\documentclass{sig-alternate-05-2015}
\usepackage{algorithm2e}

\begin{document}






%

\title{A Heterogeneous Graphical Model to Understand User-Level Sentiments in Social Media} 

%
%
%
%
%

\numberofauthors{4} 
%
\author{
%
%
\alignauthor
Rahul Radhakrishnan Iyer\\
       \affaddr{Language Technologies Institute}\\
       \affaddr{Carnegie Mellon University}\\
       \email{rahuli@andrew.cmu.edu}
\alignauthor
Jing Chen\\
       \affaddr{Language Technologies Institute}\\
       \affaddr{Carnegie Mellon University}\\
       \email{jingc1@andrew.cmu.edu}
\and
\alignauthor
Haonan Sun\\
       \affaddr{Language Technologies Institute}\\
       \affaddr{Carnegie Mellon University}\\
       \email{haonans@cs.cmu.edu}
\and  
\alignauthor
Keyang Xu\\
       \affaddr{Language Technologies Institute}\\
       \affaddr{Carnegie Mellon University}\\
       \email{keyangx@andrew.cmu.edu}
}

\maketitle


%
%

%

%
%
\printccsdesc

\begin{abstract}
Social Media has seen a tremendous growth in the last decade and is continuing to grow at a rapid pace. With such adoption, it is increasingly becoming a rich source of data for opinion mining and sentiment analysis. The detection and analysis of sentiment in social media is thus a valuable topic and attracts a lot of research efforts. Most of the earlier efforts focus on supervised learning approaches to solve this problem, which require expensive human annotations and therefore limits their practical use. In our work, we propose a semi-supervised approach to predict user-level sentiments for specific topics. We define and utilize a heterogeneous graph built from the social networks of the users with the knowledge that connected users in social networks typically share similar sentiments. Compared with the previous works, we have several novelties: (1) we incorporate the influences/authoritativeness of the users into the model, 2) we include comment-based and like-based user-user links to the graph, 3) we superimpose multiple heterogeneous graphs into one thereby allowing multiple types of links to exist between two users. 

\end{abstract}

\section{Introduction}
Online social media such as Twitter and Facebook has become pertinent way of social networking, where most of the users tend to convey personal opinions or feelings towards certain topics in their posts. Due to the large volume of online data, widespread topical variety and real-time propriety of social media, especially micro-blogging services, recent years have seen intense research work on sentiment detection in the context of online social media. Formally, sentiment detection is defined as classifying entities in social media into limited classes such as positive or negative attitudes. 

Traditional approaches focus more on supervised classification of textual information, where supervised labels are either manually labeled or using methods like distant supervision. Social text has unique characteristics such as length limit, higher probability of being noisy and the wide use of slangs and emoticons, leading to special requirements for careful consideration of data pre-processing and feature engineering. While a number of prior research work have demonstrated promising experimental results in terms of sentiment classification accuracy, most of them can hardly be extended to large-scale sentiment analysis due to their strict dependence on human annotation \cite{bermingham2010classifying} or exact matching of sentiment lexicons (e.g., SentiWordNet) \cite{esuli2006sentiwordnet}. To make things worse, even existing annotated datasets cannot catch up with the fast speed of dynamic and ever-changing social focus. It has thus become of importance to automatically better utilize the real-time social data with little or even no human intervention. The authors in \cite{tan2011user} propose a semi-supervised method that utilized metadata from user profiles to construct ground truth data (user sentiments towards a topic) and achieve promising results in predicting user level sentiments. Recently, several approaches involving machine learning and deep learning have also been used in the visual and language domains \cite{iyer2019event,iyer2019unsupervised,li2016joint,iyer2016content,iyer2018transparency,li2018object,gupta2016analysis,honke2018photorealistic,iyer2017detecting,radhakrishnan2016multiple,iyer2012optimal,qian2014parallel,iyer2017recomob}.

In this work, we propose a new semi-supervised approach that builds upon the model proposed in \cite{tan2011user}. We have further enhanced the model by: 1) incorporating the influences/authoritativeness of the users, 2) including additional user-user links involving comment-based and like-based ones, 3) superimposing the multiple heterogeneous graphs (each type of user-user link, e.g., directed t-follow, mutual t-follow, like-based etc. will have a separate heterogeneous graph as described in \cite{tan2011user}) into one thereby allowing multiple types of links to exist between two users. This has the advantage that each type of link will have a different weight and taking a weighted sum of all the types of links between two users will give better information about the ``sentiment-relatedness'' (how sentiment of one affects the other) between them. For e.g., two users having a like-based link in addition to mutual t-follow should intuitively have a stronger sentiment-relatedness when compared to two users that are just connected by mutual t-follow links.

The rest of this paper is organized as follows: Section \ref{sec:related} gives brief review of related prior work on sentiment detection in social media, where comparisons are presented. To better utilize various online user links, Section \ref{sec:proposed} proposes our new semi-supervised model, including the mathematical formulation, newly designed loss function and potential inference algorithms. Section \ref{sec:experiment} discusses potential data collection, necessary experimental settings, potential baselines, evaluation metrics and future plans. Finally, Section \ref{sec:conclusions} summarizes this paper and draw conclusions from our work. 

\section{Related Work}
\label{sec:related}
In this section, we discuss several papers that aided in the development of research on sentiment detection models in the context of social media. 

\subsection{Twitter Sentiment Classification using Distant Supervision }
While there has been a large amount of research on sentiment classification for larger pieces of text, like movie reviews~\cite{pang2002thumbs} and blogs, little has been done on classifying sentiment of messages on micro-blogging services like Twitter. Thus the paper proposes a novel approach for automatically classifying the sentiment of Twitter messages as either positive or negative with respect to a query term.


Faced with unique attributes of Twitter messages, for example, length limit of 140 characters, data availability, topical variety the high frequency of misspellings and slangs, posing challenge on classify sentiment in Twitter. For the purpose of simplicity, the sentiment can be defined as positive or negative while neutral sentiment is discarded.

In order to automatically aggregate and analyze Twitter sentiment, supervised learning framework is applied. While it is almost impossible to manually collect and label training data, the key idea is to use distant supervision. This means to use tweets with emoticons as training data, with the emoticons serving as noisy labels. Moreover, there is pre-defined set of emoticons mapped to ``:)'', indicating positive sentiment, and ``:('', indicating negative sentiment.

Moreover, characteristics of tweets have required careful design of pre-processing strategies, including (1) emoticon removal, causing classifier learn from other textual features; (2) query-term removal, avoiding the inner sentiment of query terms to bias the classification results. Further data cleaning approaches can be using equivalent class to replace strings like "@Username", "URL", and constraining consecutive repeated letters.

From Twitter Application Programming Interface (API), data are collected from April 6, 2009 to June 25, 2009, where training data is well-organized and post-processed with filters like pre-defined tweet removal strategy. Altogether there is training set, consisting of 800,000 tweets with positive emoticons and 800,000 with negative emoticons, as well as testing set, consisting of 177 negative and 182 positive tweets, all manually marked.

Based on features like unigrams, bigrams, unigrams and bigrams, and parts of speech (POS), comparisons have been performed between the keyword baseline model and machine learning classifiers like Naive Bayes (NB), Maximum Entropy (MaxEnt) and Support Vector Machine (SVM). Experiments have shown that all classifiers have greatly improve over the keyword baseline, reaching over 80\% accuracy when trained with emoticon data. More specifically, while classifiers with Unigram features see similar performance with previous work~\cite{pang2002thumbs}, Bigram features aims to capture phrases like ``not good'', but alone is not performing as expected due to sparse feature space. Combining Unigram and Bigram features see the most promising gains and indicates the potential of automated classifiers on Twitter sentiment detection.

In general, this work can be viewed as an empirical study of how supervised classification can be applied in Twitter sentiment analysis with careful design of pre-processing strategy. There is still much that could be taken into considerations, including utilizing semantics to improve classification performance, better handling domain-specific tweets and extend English Twitter sentiment classification to other languages apart from English.



\subsection{Sentiment Analysis of Twitter Data} 
\label{subsec:sentiment-apoorv}
Apoorv et all also examined sentiment analysis on Twitter data and they have two main contributions: (1) introducing POS-specific prior polarity features; (2) exploring the use of a tree kernel to obviate the need for tedious feature engineering. 

A number of their features are based on prior polarity of words. To obtain the prior polarity of words, they use Dictionary of Affect in Language (DAL) \cite{whissel1989dictionary} and extend it using WordNet. DAL contains about 8000 English language words each of which are assigned a pleasantness score between 1 (Negative) - 3 (Positive). They first normalize the scores by diving each score by the scale (which is equal to 3). Then words with polarity less than 0.5 are considered as negative, words with higher than 0.8 are considered as positive and the rest as neutral. For a word which cannot be directly found in DAL, if any synonym retrieved from Wordnet is found in DAL, the word is  assigned the same pleasantness score as its synonym and if not, the word is not associated with any prior polarity. 

Beside prior polarity of words, they design a tree representation of tweets to combine many categories of features. To calculate the similarity between two trees they use a Partial Tree (PT) kernel first proposed by Moschitti \cite{moschitti2006efficient}. A PT kernel calculates the similarity between two trees by comparing all possible sub-trees. This calculation is made computationally efficient by using Dynamic Programming techniques. By considering all possible combinations of fragments, tree kernels capture any possible correlation between features and categories of features. In this manner, it is not necessary tocreate by hand features at all levels of abstraction.

The advantage of this work is that it take POS-specific polarity scores into consideration but other linguistic analysis like parsing, semantic analysis and topic modeling are not considered. Besides, the machine learning classifier they use is SVMs and SVM cannot potentilally reach the state-of-art performance. Tring using better and more proper machine learning models might be another way of making breakthrough based on this work.

\subsection{User-level sentiment analysis incorporating social networks}
\label{subsec:user-sentiment}
Tan et al. proposed \cite{tan2011user}  a semi-supervised method to detect user level sentiment on social media. The motivation is that users on social networks are connected with each other and those connected users are more likely to share common opinions. Also, some meta data created by users themselves can serve as the ground truth opinion towards specific topics. 

In order to classify each users' sentiment on specific topics into binary polarities, "positive" and "negative", some users who explicitly labeled their sentiments are chosen as root users. Then  four possible connections are used to expand the graph, including mutual and directed follow links and mutual and directed retweet links. After adding topic related tweets to selected users, a directed heterogeneous graph on one topic is constructed. Given ground truth labels of part of users, the target is to estimate the sentiments of other users which maximize the log likelihood functions as follows: 
\begin{equation}
\resizebox{0.43\textwidth}{!}{$
\begin{split}
\log P(\mathbf{Y}) &= (\sum_{v_i \in V} [ \sum_{t \in {tweets_{v_i, k, l}}}{\mu_{k,l} f_{k,l}(y_{v_i},\hat{y_t})}\\
& + \sum_{v_j \in {Neighbors_{v_j, k, l}}}{\lambda_{k, l} h_{k,l}(y_{v_i},y_{v_j})}  ]) \\
& -\log Z
\end{split}
$}
\end{equation}

where $Z$ is the normalization factor and the first two inter sum indicate user-tweet factors and user-user factors respectively and indices $k,l$ range over sentiments labels $\{0,1\}$. $f_{k,l}(\cdot,\cdot)$ and $h_{k,l}(\cdot,\cdot)$ are features function, and $\mu_{k,l}$ and $\lambda_{k,l}$ are parameters representing feature impact. Thus we use $w_{labeled}$ and $w_{unlabeled}$ to indicate different levels of confidence in users that were or were not initially labeled as follow:
\begin{equation}
\resizebox{0.43\textwidth}{!}{$
f_{k,l}(y_{v_i},\hat{y_t})=
    \begin{cases}
      \frac{w_{labeled}}{|tweets_{v_i}|}, & y_{v_i}=k,\hat{y_t}=l, v_i \text{ is labeled} \\
      \frac{w_{relation}}{|tweets_{v_i}|}, & y_{v_i}=k,\hat{y_t}=l, v_i  \text{ is unlabeled} \\
      0, & \text{otherwise} 
    \end{cases}
$}
\end{equation}

\begin{equation}
h_{k,l}(y_{v_i},y_{v_j})=
    \begin{cases}
      \frac{w_{relation}}{|Neighbors_{v_i}|}, & y_{v_i}=k,y_{v_j}=l \\
      0 & \text{otherwise}
    \end{cases}
\end{equation}
Need to note that the values of those weights are usually tuned heuristically. For example, initially-labeled users receive highest confidence. The setting of this work selects $w_{labled} = 1.0$, $w_{unlabeled}=0.125$, $w_{relation}=0.6$. 

The authors then utilize two methods to estimate the parameters $\lambda_{k,l}$ and $\mu_{k,l}$ in log-likelihood function, including direct estimation from simple statistics and SampleRank\cite{rohanimanesh2011samplerank}. Then \textbf{Loopy Belief Propagation} (LBP) is utilized to perform inference for the given model. Need mention due to the randomization of SampleRank, this paper utilize majority votes of 5 predictions to label each user. 

Three user-classification methods are compared, including Majority-vote using SVM and two proposed method using different estimating approaches. Results show proposed models perform significantly better than the baseline. Also, This paper shows that follow links are a stronger agreement than retweet links and directed links perform better than undirected links. However, two parameter estimating methods show similar performances. 

While this paper produces promising results in classifying user-level sentiments, it make some assumptions to simply the problem. For example, it only consider user importance on aspect of labeled or not, ignoring its connection information in the heterogeneous graph. 
Also, no text information is used in this work, which has been proved to be a powerful indicator of sentiments.

\section{Proposed Model}
\label{sec:proposed}
We propose a model that is built on top of \cite{tan2011user}, with several enhancements. This model put forth by the authors has numerous advantages, for e.g., 1) it being semi-supervised and requiring little no human annotation, 2) exploiting the network structure in social media, which takes a turn from the usual sentiment analysis approaches that are all at a text-level, 3) the sentiments are calculated at a user-topic level. However, there are still several drawbacks that can be addressed.

Firstly we note that the influence/authoritativeness of the users is not considered at all in the model. Influential users have more tendency to affect (for better or worse) the sentiment of the users connected to them. So, this is an important part of the graphical model that has not been considered. Secondly, the model proposed by the authors consider four different heterogeneous graphs, one for each type of user-user link, e.g., mutual t-follow, directed t-follow etc. But this is not the best representation because there can exist multiple types of links between two users and a weighted average of all these links can give a better prediction of the sentiment-relatedness between them. For e.g., two users having a like-based link in addition to mutual t-follow should intuitively have a stronger sentiment-relatedness when compared to two users that are just connected by mutual t-follow links. Thirdly, several other user-user links like comment-based links and like-based links are not considered. This can further strengthen the graphical model.

With all of this in mind, we define a newer and enriched graphical model that addresses all these shortcomings. The different enhancements are given below:
\begin{enumerate}
\item We compute a user's influence in the network as his/her \textbf{pagerank} score in the graph (considering only the user-user links in the heterogeneous graph: removing the user-tweet links) considered.
\item We allow multiple types of links to exist between two users simultaneously, which in essence is a superimposition of all the different heterogeneous graphs (one heterogeneous graph for each type of user-user link) into one graph. For e.g., there could be a retweet link, t-follow link and a like-based link between two users and a combination of these can be more predictive of the sentiment-relatedness between them.
\item We introduce four new user-user links: 1) directed like-based links (one user likes the posts of another but the reverse need not be true), 2) mutual like-based links (both users like each other's posts), 3) directed comment-based links (one user comments on the tweets of another but the reverse need not be true), 4) mutual comment-based links (both users comment on each other's tweets). This will enrich and enhance the graphical model as all these factors: like, comment, retweet, t-follow, are all indicative of shared sentiment between two users.
\end{enumerate}

\begin{figure}
\label{fig:graph_model}
\includegraphics[scale=0.19]{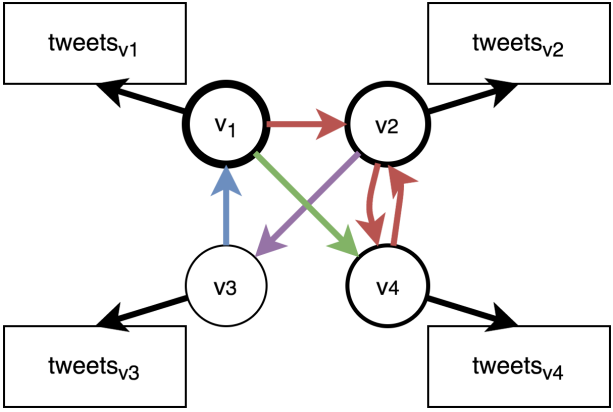}
\caption{An example of proposed directed heterogeneous graph on a specific topic. Edges in different colors indicate different types of relation dependencies. The thickness of user circle $v$ shows its influence in the graph.}
\end{figure}

The proposed graphical model is shown in Figure 1, where we incorporate textual, user and network information into a single heterogeneous graph for a specific topic. We define $G$ to be the set of different types of user-user links. In essence, $G$ will consist of $8$ different types of links: the $4$ old ones (mutual \& directed t-follow, mutual \& directed retweet) that are described in \cite{tan2011user} and the $4$ new ones described above. In our new model, which we will describe below, since we consider mutliple types of links to exist between two users in the same graph, we will need to partition (with overlap, not disjoint) the neighbours of a user according to $G$. Let $v_i$ be the root user considered and let $V$ be the set of neighbor users of user $v_i$. Formally, we define the partition function, \textsc{Partition}:$V, G, v_i \mapsto 2^V$, as follows:
\begin{equation}
\resizebox{0.43\textwidth}{!}{$
\textsc{Partition}(V, G, v_i) = \lbrace \lbrace v_j\:|\: v_j \in V, \; (v_i, v_j) \; \text{is a type of link} \in G \rbrace \rbrace
$}
\end{equation}

Basically what this function does is: given a neighbor set of users for a root user, we split that set into multiple smaller sets (with overlap, not disjoint) such that each of those sets corresponds to those neighbors that share the same type of link with the root. These types of links are in $G$.

Now, we assume that we considering a particular topic $T$ to make things easy. The problem then is to determine the sentiments of the users towards this particular topic. Let the set of users be denoted by $V = \lbrace v_i \rbrace$. Let $y_{v_i}$ be the label for user $v_i$, and \textbf{Y} be the vector of labels for all users. We make the markov assumption, as in \cite{tan2011user}, that the user sentiment $y_{v_i}$ is influenced only by the (unknown) sentiment labels $\hat{y_t}$ of tweets $t \in tweets_{v_i}$ (the tweets of the user $v_i$) and the probably unknown sentiment labels of the immediate user neighbors $Neighbors_{v_i}$ of $v_i$. Thus, with all of this, we can now propose our new factor-graph-based model as:
\begin{equation}
\resizebox{0.43\textwidth}{!}{$
\begin{split}
\log P(\mathbf{Y}) &= ( \sum_{v_i \in V} p_{v_i}[\sum_{t \in tweets_{v_i};\:k,l} \mu_{k,l}f_{k,l}(y_{v_i}, \hat{y_t})\\
 & + \sum_{g \in \textsc{Partition}(Neighbors_{v_i}, G, v_i)}\sum_{v_j \in g;\:k,l} \lambda_{k,l,g} h_{k,l,g}(y_{v_i},y_{v_j})])\\
 & - \log Z
\end{split}
$}
\end{equation}

where the indices $k,l$ range over the set of sentiment labels $\lbrace 0,1 \rbrace$, and $g \in G$. $Z$ is the normalization factor. $f_{k,l}(\cdot,\cdot)$ and $h_{k,l,g}(\cdot, \cdot)$ are feature functions which we define below, and $\mu_{k,l}$ and $\lambda_{k,l,g}$ are parameters that determine the ``weight/impact'' of that \textit{category} of link. Here, this category of links is not that same as the \textit{types} of links discussed earlier. For e.g., $\lambda_{0,1,\texttt{mutual t-follow}}$, essentially denotes the impact of the mutual t-follow user-user link between two users where the first user has a negative sentiment towards the topic in consideration (hence the $0$) and the second user has a positive sentiment (hence the $1$). Similarly, $\mu_{1,0}$ represents the impact of that type of user-tweet link where the user's overall sentiment towards the topic (essentially his label $y_{v_i}$) is positive (hence the $1$) but his tweet $t$ is negative (hence the $0$). These impact/weights are parameters that are learnt by the model. It could happen that the model learns a very small value for $\mu_{0,1}$ because such links are rare or don't have much impact. It is important to note that these parameters are global and are not specific to users. With the new model, we have more number of parameters to estimate than the original model: it is easy to see that we need to estimate $4$ parameters to determine $\mu_{k,l}, \forall k,l \in \lbrace 0,1 \rbrace$ and $4 \times |G|$ to determine $\lambda_k,l,g, \forall k,l \in \lbrace 0,1 \rbrace$ and $g \in G$.

Here, $p_{v_i}$ is the \textit{pagerank} score of user $v_i$ in the graph. As before, the first and second inner sums correspond to user-tweets factors and user-user factors respectively. The improvements in the proposed model are immediately visible. The pagerank score $p_{v_i}$ before the first summation takes into account the influence of the users $v_i \in V$ to determine the sentiment-label for the users. In addition, we see that in the second sum, we partition the neighbor space of $v_i$ according to the function \textsc{Partition} defined earlier. This way, we get a component for each type of user-user link and in consequence, we learn a different weight $\lambda_{k,l,g}$ for each category and type of link. Thus, each type of user-user link gets its own weight and a combination of all these links will give a better representation of the sentiment-relatedness between two users.

As in the original model proposed by the authors, explained above, the feature functions $h_{k,l,g}$ and $f_{k,l}$ are defined below:

\begin{equation}
\resizebox{0.43\textwidth}{!}{$
f_{k,l}(y_{v_i},\hat{y_t})=
    \begin{cases}
      \frac{w_{labeled}}{|tweets_{v_i}|}, & y_{v_i}=k,\hat{y_t}=l, v_i \text{ is labeled} \\
      \frac{w_{relation}}{|tweets_{v_i}|}, & y_{v_i}=k,\hat{y_t}=l, v_i  \text{ is unlabeled} \\
      0, & \text{otherwise} 
    \end{cases}
$}
\end{equation}

\begin{equation}
h_{k,l,g}(y_{v_i},y_{v_j})=
    \begin{cases}
      \frac{w_{relation}}{|Neighbors_{v_i}(g)|}, & y_{v_i}=k,y_{v_j}=l \\
      0, & \text{otherwise}
    \end{cases}
\end{equation}

Here, $g$ indicates the group of links that are defined in $G$. The estimation procedures are similar to the original approach:
\begin{enumerate}
\item \textbf{Direct Estimation:} From the labeled subset of data, albeit limited, one can use direct statistical estimation techniques such as simple counts. Let $E_{labeled}$ be the subset of edges in our heterogeneous graph in which both the end-points are labeled. We estimate the $4 \times |G|$ user-user parameters as follows:
\begin{equation}
\resizebox{0.43\textwidth}{!}{$
	\lambda_{k,l,g} = \frac{\sum_{(v_i,v_j) \in E_{labeled}} I(y_{v_i} = k, y_{v_j} = l, G = g)}{\sum_{(v_i,v_j) \in E_{labeled}} I(y_{v_i} = k, y_{v_j} = 1, G = g) + I(y_{v_i} = k, y_{v_j} = 0, G = g)}
    $}
\end{equation}

where $I(\cdot)$ is the indicator function. It is important to note that we do not have the labels for individual tweets in the dataset (that is the essence of the semi-supervised learning approach of this paper). We therefore make the strong assumption that positive users only post positive tweets on the topic, and negative users post only negative tweets. In essence, we set $\mu_{k,l} = 1$ if $k = l$, 0 otherwise.

\item \textbf{SampleRank}: Here, instead of using direct statistical estimation techniques to esimate the parameters, we \textit{learn} the parameters. This algorithm is described in Algorithm \ref{algo:samplerank}.
\end{enumerate}

As in the original paper, we perform inference using \textbf{loopy belief propagation}.

\section{Experiments Design}
\label{sec:experiment}
\subsection{Data Collections}
\label{subsec:data}
Our dataset can be extracted from social media website that contains both user-user links and user-content links. Twitter is a good option, which is used in previous studies.\cite{tan2011user} Similar to this work, we can first collect a large dataset from twitter and then select several topics that covers multiple aspects. For examples, we should consider both frequent topics and rare queries. Then for each topic, we can generate root users set as ground truth data by analyzing metadata or labels created by users themselves. Then we can use user-user connections to expand users set. In previous work, the size of root user set is less than one thousand and expanded graph usually contains over one million users and tweets as vertexes. Since previous work is done five years ago, those topics and posts might be outdated. We prefer to collect new data instead of using the old one. Also we might examine the size of data for each selected topics to decided whether to collect more data or change topics. On the other hand, we also have to collect extra connections information such as like or comments to train different parameters for links. 

\subsection{Implementation}
\label{subsec:implementation}
PageRank and some text classification algorithm should be implemented to get user importance and text information for baseline approaches respectively. As for features for text classification method, we can try bag of word or some more sophisticated features such as ngram or word2vec.  

Furthermore, the methods to do parameters estimation and inference can be the same as previous work\cite{tan2011user}, including SampleRank and loopy belief propagation. The SampleRank algorithm is stated in Algorithm \ref{algo:samplerank}.

$RelPerf(\mathbf{Y}^{new},\mathbf{Y})$ is the difference in performance, measured on the labeled data only, between $\mathbf{Y}^{new}$ and $\mathbf{Y}$, where the performance $Perf(\mathbf{Y}) = Accuracy(\mathbf{Y}) + MacroF1(\mathbf{Y})$.  $LLR_{\phi}(\mathbf{Y}^{new}, \mathbf{Y})$ is the log-likelihood ratio for the new sample $\mathbf{Y}^{new}$ and the previous label set \textbf{Y}: $LLR_{\phi}(\mathbf{Y}^{new}, \mathbf{Y}) = \log (\frac{P(\mathbf{Y}^{new})}{P(\textbf{Y})})$.

\begin{algorithm}
\label{algo:samplerank}
\SetKwData{Left}{left}\SetKwData{This}{this}\SetKwData{Up}{up}
\SetKwFunction{Union}{Union}\SetKwFunction{FindCompress}{FindCompress}
\SetKwInOut{Input}{input}\SetKwInOut{Output}{output}
\Input{Heterogeneous graph HG with labels on some of the user nodes}
\Output{Parameters values $\phi$ and full label-vector \textbf{Y} }
\BlankLine
Randomly initialize \textbf{Y};\\
Initialize $\phi$ from NoLearning;\\
\For{$i:= 2$ \KwTo $\textrm{Number of Steps}$}{
$\mathbf{Y}^{new} := \text{Sample}(\mathbf{Y})$ \;

\If(){(RelPerf$(\mathbf{Y}^{new},\mathbf{Y})>0$ \text{and} $LLR_{\phi}(\mathbf{Y}^{new},\mathbf{Y})<0)$  \\ \tcp{peformance is better but the objective function is lower} \textbf{or} $(RelPerf(\mathbf{Y}^{new},\mathbf{Y})<0$ \text{and} $LLR_{\phi}(\mathbf{Y}^{new},\mathbf{Y})>0)$ \\ \tcp{peformance is worse but the objective function is better} }{
$\phi := \phi - \eta \Delta_{\phi}{LLR_{\phi}(\mathbf{Y}^{new},\mathbf{Y})}$\;
}

\If{converge}{break\;} 
\If{$RelPerf(\mathbf{Y}^{new},\mathbf{Y}) >0$ }{$\mathbf{Y}:= \mathbf{Y}^{new}$\;} 
}
\caption{SampleRank algorithm}
\end{algorithm}

\subsection{Baselines}
In this section, we discuss some potential baselines to compare with our method. 
\begin{itemize}
\item \textbf{Text-based SVM majority vote}. After training a SVM classifier\cite{fan2008liblinear}, we can predict each tweet in the heterogeneous graph. Then we make predictions for users based on majority vote of their posted tweets.

\item \textbf{Text-based SVM batch prediction} Instead of using majority vote for multiple tweets, we feed those tweet as a single bag of word document to the SVM. In this case, for a specific user, sentiment is predicted based on all his posts combined in a batch. 

\item \textbf{Tan et al. work}. This is the method we discussed in Section 2.3. \\

As for the training set for tweet-level SVM, we can adopt some available datasets constructed by previous work, such as \cite{read2005using}. In case of those datasets are outdated due to the fast updating pace of social media, we can also construct our own topical related training set with the help of crowdsourcing platform such as MTurk.

\end{itemize}
\subsection{Evaluations}
As for evaluation metrics, we choose to use accuracy and MacroF1 as following:
\begin{equation}
\text{Accuracy} = \frac{\#TP + \#TN}{ \#TP + \#TN + \#FN + \#FP}
\end{equation}

\begin{equation}
F1 = \frac{2}{\frac{\#TP}{\#TP+\#FP}+\frac{\#TP}{\#TP+\#FN}}
\end{equation}
\begin{equation}
\text{MacroF1} = \frac{F1_{pos} + F1_{neg}}{2}
\end{equation}
where $\#TP$, $\#TN$, $\#FP$, $\#FN$ indicate the numbers of true positive, true negative, false positive and false negative predictions respectively.

The ground truth data for users can be collected with the help of human annotations.

\section{Conclusions}
\label{sec:conclusions}
To summarize, this work has investigated sentiment classification in the context of social media, more specifically, on Twitter data. We first present a detailed literature review of related work and introduce the motivation of our work. We then propose a semi-supervised model and corresponding inference algorithm to predict user-level topical sentiment on Twitter data. Compared with previous work, our new model takes into consideration various user-link information in the web heterogeneous graph, including the impact of different users, and differences among various kinds of links. Lastly, we briefly discuss potential data collection, necessary experimental settings and evaluations.

%
\bibliographystyle{abbrv}
\bibliography{sigproc}  
%
%
\end{document}